\pdfoutput=1

\documentclass[11pt]{article}

\usepackage[final]{acl}

\usepackage{amsmath}
\usepackage{times}
\usepackage{latexsym}
\usepackage{mathabx}
\usepackage{algorithm}
\usepackage{algpseudocode}

\usepackage[T1]{fontenc}

\usepackage[utf8]{inputenc}

\usepackage{microtype}

\usepackage{makecell}
\usepackage{booktabs}
\usepackage{graphicx}
\usepackage{multirow}
\usepackage[labelfont=bf, format=plain, labelformat=simple, font=small, compatibility=false]{caption}

\title{Label Agnostic Pre-training for Zero-shot Text Classification}

\author{
  Christopher Clarke\hspace{10pt} Yuzhao Heng\hspace{10pt} Yiping Kang\hspace{10pt} Krisztian Flautner\hspace{10pt} \\  \textbf{Lingjia Tang\hspace{10pt}  Jason Mars\hspace{10pt}}    \vspace{0.3cm}\\
  \text{Computer Science \& Engineering} \\
    \text{University of Michigan} \\ \text{Ann Arbor, MI}\\
    \text{\{csclarke, stefanhg, ypkang, manowar, lingjia, profmars\}@umich.edu} \\
}

\begin{document}
\maketitle
\begin{abstract}
Conventional approaches to text classification typically assume the existence of a fixed set of predefined labels to which a given text can be classified. However, in real-world applications, 
there exists an infinite label space for describing a given text. 
In addition, depending on the aspect (sentiment, topic, etc.) and domain of the text (finance, legal, etc.), the interpretation of the label can vary greatly. 
This makes the task of text classification, particularly in the zero-shot scenario, extremely challenging. In this paper, we investigate the task of zero-shot text classification with the aim of improving the ability of pre-trained language models (PLMs) to generalize to both seen and unseen data across varying aspects and domains. 
To solve this we introduce two new simple yet effective pre-training strategies, \textit{Implicit} and \textit{Explicit pre-training}. These methods inject aspect-level understanding into the model at train time with the goal of conditioning the model to build task-level understanding. To evaluate this, we construct and release UTCD, a new benchmark dataset for evaluating text classification in zero-shot settings.
Experimental results on UTCD show that our approach achieves improved zero-shot generalization on a suite of challenging datasets across an array of zero-shot formalizations.

\end{abstract}

\section{Introduction} \label{intro}
\begin{figure}
  \centering
    \includegraphics[width=1\columnwidth]{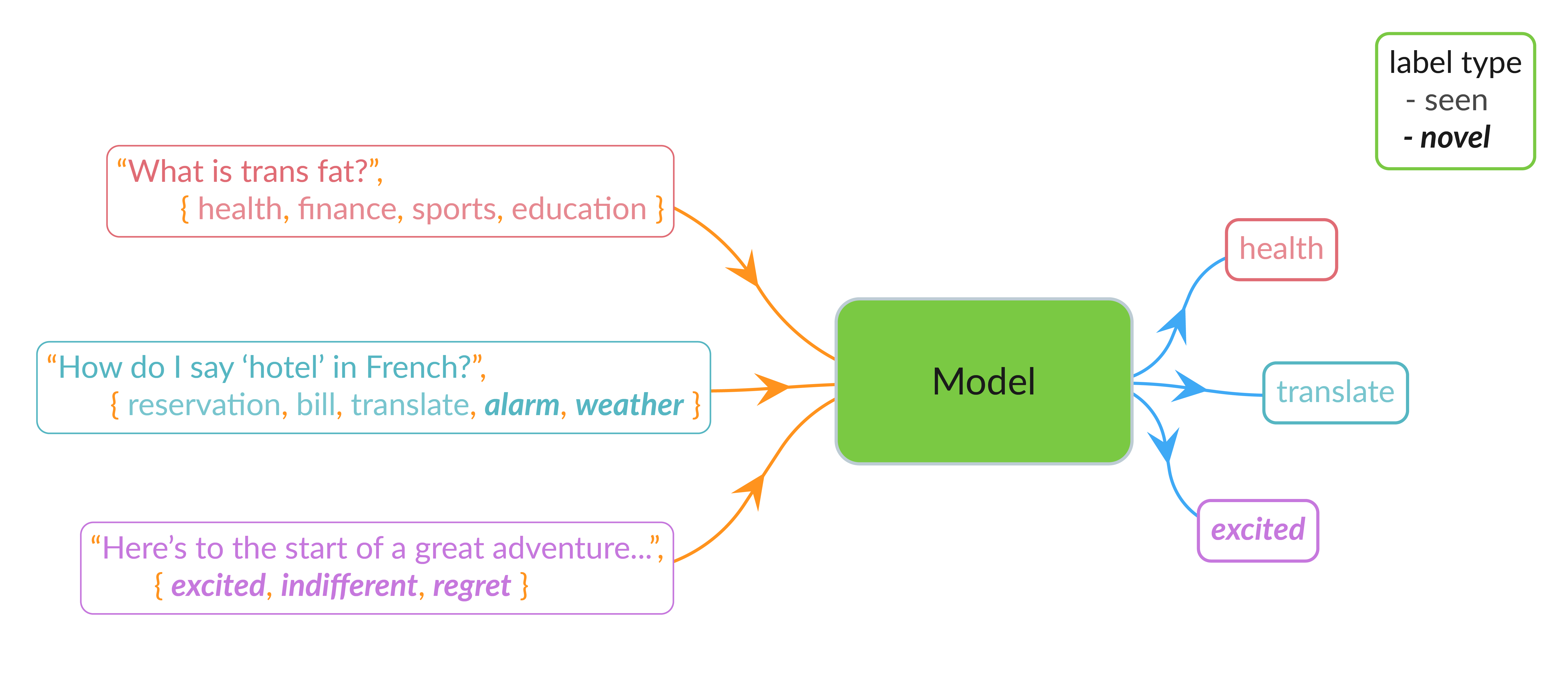}
    \caption{Zero-shot Text Classification Problem: In real-world applications, the model needs to adapt to unseen labels. For a given aspect and domain, the interpretation of a given text-label pair can vary greatly.}
  \label{fig:problem}
  \vspace{-1pc}
\end{figure}

Text classification is the process of categorizing text into sets of organized groups where each set consists of similar content in a well-defined manner \cite{txt-cls, joulin2016bag}. 
Supervised approaches have achieved great success in recent years due to the availability of rich training data and the advent of large pre-trained language models such as BERT \cite{bert}. These conventional approaches typically assume the presence of a pre-defined set of labels to which a given text can be classified. 
However, in real-world applications, several challenges emerge: 

\textbf{1)} The label space is constantly evolving. Over time, new labels are constantly emerging and the definition of the label space is constantly being refined. 
For example, intent classification systems such as those used in chatbots and dialogue systems are constantly introducing new intents as their range of supported features increases. Social networks such as Twitter encounter new and emerging topics on a daily basis from massive amounts of content that need to be classified. Figure \ref{fig:problem} shows an example of this emerging label space.

\textbf{2)} The range of applications for text classification is vast. Text classification is pivotal to many different application areas from sentiment analysis to topic labeling, etc, and is used in a variety of domains such as finance, health, etc. When applied to this conglomeration of uses, it is typically assumed that there exists a comprehensive dataset of well-defined text-label pairs for each use case. However, in many real-world settings, annotated data is either scarce or unavailable entirely. Additionally, the use of dedicated models for each task is impractical due to the additional compute overhead and maintenance, thus making it difficult to scale over time.

Zero-shot learning (ZSL) is aimed at addressing these constraints. Zero-shot Learners are models capable of predicting unseen classes. When applied to text classification, these models aim to associate a piece of text with a given label without the need for having been trained on that label. However, despite recent advancements in the capabilities of PLMs, zero-shot models still vastly underperform their supervised counterparts \cite{train-once, gpt2-nvidia, gpt3}. As such, this remains an open research problem. 

In this paper, we investigate the challenge of reducing the aforementioned performance gap present in these zero-shot models compared to their supervised counterparts on unseen data. We theorize that the poor generalization of these zero-shot models is due to their lack of aspect-level understanding during their training process. To alleviate this we introduce two new simple yet effective pre-training strategies, \textit{Implicit} and \textit{Explicit pre-training} which specifically inject aspect-level understanding into the model.

In order to evaluate these strategies, we canvas the range of zero-shot formalizations for enabling zero-shot text classification on PLMs and apply our techniques. Additionally, we introduce the Universal Text Classification Dataset (UTCD), a large-scale text classification dataset for evaluating zero-shot text classification. UTCD is a compilation of 18 classification datasets spanning 3 main aspects of Sentiment, Intent/Dialogue, and Topic classification. Our results on UTCD show that by employing both our implicit and explicit pre-training strategies we can achieve improved zero-shot performance on a suite of challenging datasets for which the model was not trained on.

Specifically, this paper makes the following contributions:

\begin{itemize}
     
    \item We introduce \textit{Implicit \& Explicit} pre-training, two new simple yet effective pre-training strategies for improving zero-shot performance.
     
    \item We construct and release UTCD, a new benchmark dataset for evaluating text classification systems across a suite of diverse tasks and domains. We release our models and dataset\footnote{\url{https://github.com/ChrisIsKing/zero-shot-text-classification}}.
     
    \item We conduct a thorough evaluation of various zero-shot text classification formalizations showing the effectiveness of our training strategies on each as well as insights gained.

\end{itemize}

\begin{figure*}
  \centering
  \includegraphics[width=1\textwidth]{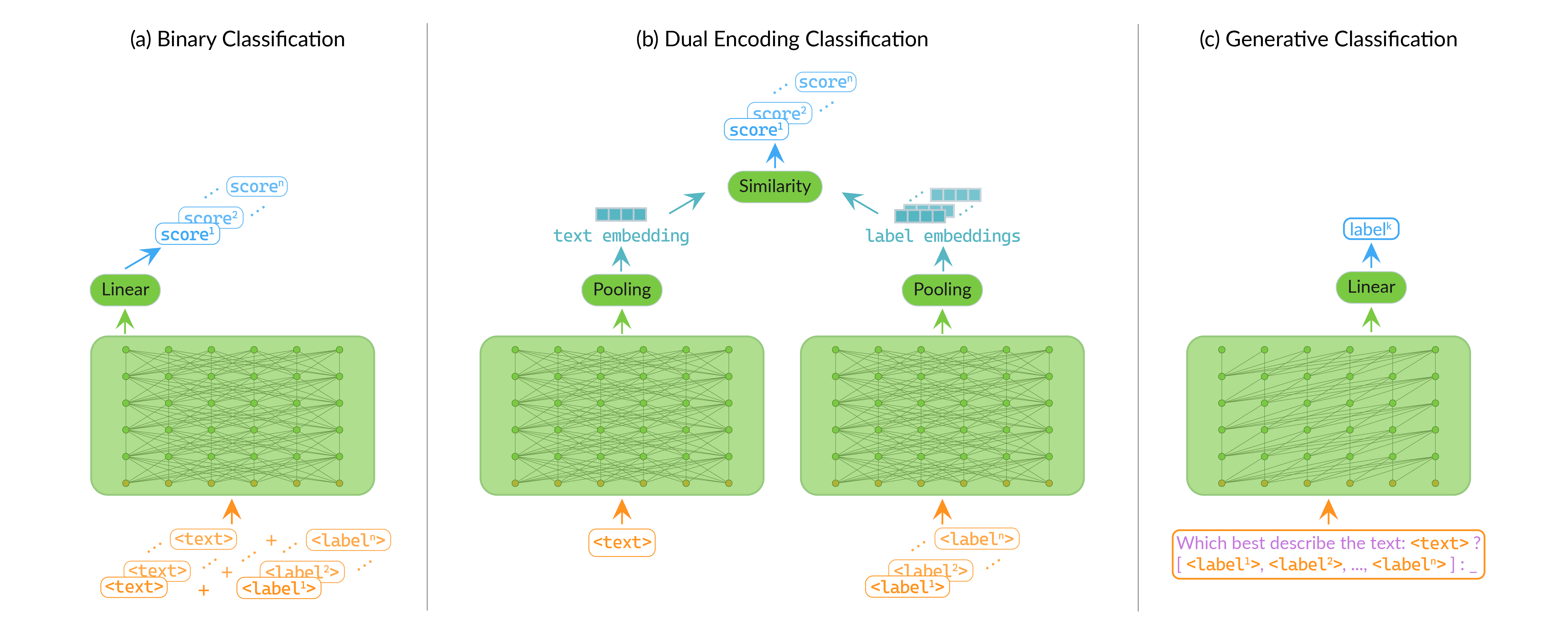}
  \caption{Zero-shot Text Classification Formalizations: \textbf{Part (a)} illustrates the binary classification formalization described in section \ref{fig:formalization} where concatenated <text, label> pairs are passed as input to the model. \textbf{Part (b)} illustrates dual encoding where text label pairs are encoded separately and scored via a distance metric. \textbf{Part (c)} illustrates text classification where the model generates desired label based on a natural language instruction template.}
  \label{fig:formalization}
   
\end{figure*}

\section{Task Formulation} \label{sec:formulation}
\newcommand{\argmax}[1]{\underset{#1}{\operatorname{arg}\operatorname{max}}\;}

In this section, we introduce the task of zero-shot text classification and describe a set of formalizations for facilitating the classification of text in a zero-shot manner, i.e. being able to predict unseen labels.

\paragraph{Conventional Text Classification} Text classification approaches using PLMs assume the existence of a pre-defined set of labels $\left\{y_i\right\}_n^1$ where for a given input sequence $X$, the model outputs a representation of that sequence as a sequence of hidden states $\left\{h_i\right\}_l^1$. 
Hidden states in the final layer are pooled to a single vector. In the case of BERT \cite{bert}, the $[\mathrm{CLS}]$ token is taken, and a linear softmax layer is added to predict the probability distribution of the label set:

\begin{equation}
    \vec{\mathrm{P}} \left( \left\{y_i\right\}_n^1 \big | ~h \right) = \mathrm{softmax}(W h)
\end{equation}

For the zero-shot scenario, this approach breaks since the output class set $\left\{y_i\right\}_n^1$ is fixed. This prevents the classification of text to new labels unless the model is re-trained with the new label set or a mapping of existing labels to unseen labels is built, both of which are impractical and cumbersome for real-world scenarios.

\subsection{Binary Zero-shot Classification}
To facilitate zero-shot classification of PLMs, ~\citet{tars, train-once, benchmark&bert-nli} formulate text classification as a series of binary classification tasks:

\begin{equation}
    f(\mathrm{label}(y_i), x) = \mathrm{P} \left( \mathrm{True} ~|~ y_i, x \right) \
\end{equation}

The model is provided with a concatenation of the class label $\mathrm{label}(y_i)$ and input text and the output layer generates a binary $\mathrm{True}/\mathrm{False}$ prediction with a confidence score $\mathrm{P}$. The $\mathrm{True}$-prediction class with the highest confidence is selected as the final prediction, that is, 

\begin{equation}
    \hat{y} = \argmax{i \;\in\; \{1 \ldots n\}} f(\mathrm{label}(y_i), x)
\end{equation}
where $n$ is the number of classes/labels. Such cross-attention (CA) models apply attention layers on the text and labels jointly, which intuitively allows for rich interactions. This architecture is shown in part (a) of Figure \ref{fig:formalization}.

\subsection{Dual Encoding Zero-shot Classification}
In contrast to cross-attention based architectures, Dual Encoder models \citep{sbert, dual-encoder, clarke-etal-2022-one} instead focus on learning representations for a given text and label independently. They separately embed the text and label, via an encoder $\Phi$ and compute pair-wise scores $S$ based on the encoded representations with a distance metric $Dist$, such as dot-product or cosine similarity: 

\begin{equation}
    S(x, y_i) = Dist \left( \Phi(x), \Phi(y_i) \right)
\end{equation}
Sentence-Bert \citep{sbert} takes PLMs such as BERT and RoBERTa as the base encoder and use siamese networks to derive sentence embeddings by comparing similarities between sentence pairs as shown in part (b) of Figure \ref{fig:formalization}. For text classification, this architecture can be used to derive latent representations for a given text and label and classify a sequence $x$ according to:

\begin{equation}
    \hat{y} = \argmax{i \;\in\; \{1 \ldots n\}} S(x, y_i)
\end{equation}

\subsection{Generative Classification}
Lastly, the generative formulation of zero-shot text classification uses autoregressive language models by passing in text and label sets as natural language prompts and training the model to generate the target label token by token. 
As described in \citet{gpt2-nvidia}, we reformulate the text classification problem as a multiple choice question answering problem. The model is provided with a multiple-choice question description containing each class label in natural language, and trained to generate the correct answer, as shown in part (c) of Figure \ref{fig:formalization}. The intuition behind this approach is to train the model to use common sense reasoning to select the most probable description of the text data from a provided list of rich natural language classes. Given some input text $t$, the model is optimized with the next token prediction language modeling loss:

\begin{equation}
    \sum_t \mathcal{L}(w_t, P(\hat{w}_t|w_{[1,t-1]}))
\end{equation}

\begin{figure*}[t]
    \centering
    \includegraphics[width=1\textwidth]{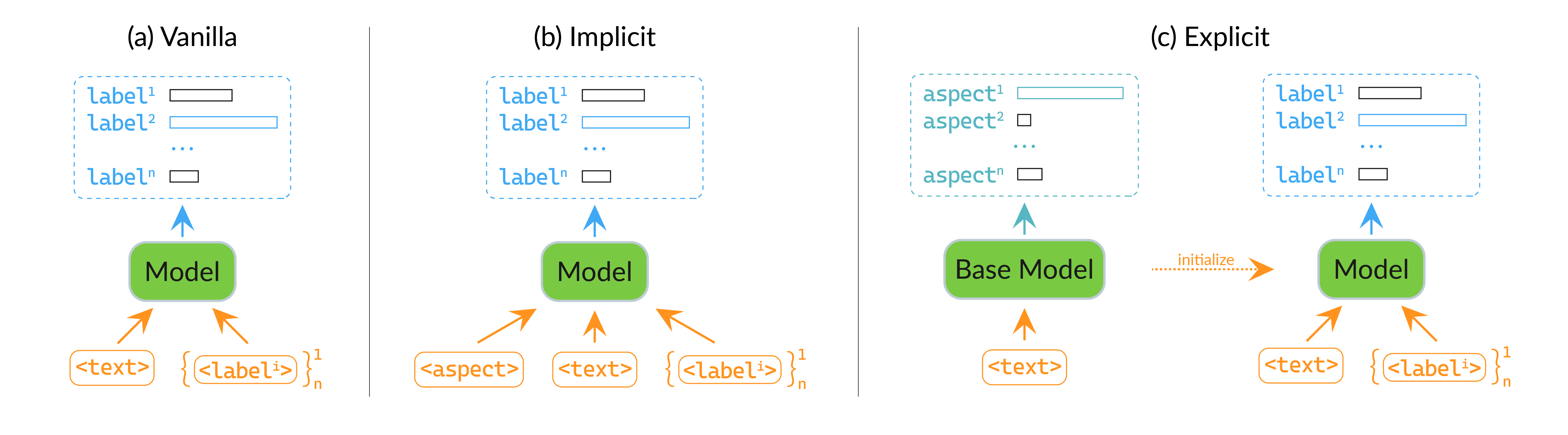}
    \caption{Zero-shot Text Classification Training Strategies. \textbf{Part (a)} shows standard model training where a text and the set of label options are passed to the model. \textbf{Part (b)} illustrates implicit training where the aspect is additionally passed as input. \textbf{Part (c)} shows injecting aspect knowledge to the model explicitly through gradient update, to initialize subsequent training. }
    \label{fig:train-strategies}
    
\end{figure*}

\section{Method} \label{approach}
In this section, we outline the methodology for our \textit{Implicit} \& \textit{Explicit} pre-training strategies which allow us to inject aspect-specific knowledge into PLMs to improve generalization to unseen data. We first define the term aspect and outline the gap between the performance of the zero-shot models shown in section \ref{fig:formalization} on seen data compared to that of unseen data. Lastly, we describe our intuition behind why localization of aspect knowledge helps to bridge this gap.

\paragraph{Aspect Definition} In the scope of this work, we define an aspect as the type of task to which a given set of datasets belong too. For example, sentiment is considered an aspect because it cleanly defines a task definition of understanding the emotion conveyed in a given text. This definition holds true even if the domain of the data changes. e.g sentiment detection of news data vs sentiment of social media tweets. In addition to having a clean task definition, we stipulate that the set of labels considered in a given aspect must convey that aspect. e.g For intent, the label \textit{"turn off alarm"} conveys that the text describes the intention to do something.

\subsection{Transfer Learning for Text Classification}
The prevailing method for training models to perform classification tasks is to add a linear head on top of a pre-trained language model and fine-tune the entire network on labeled data \cite{bert}. However, when scaled to multi-task, multi-domain applications these models suffer from issues such as catastrophic forgetting and conflicting knowledge transfer across tasks \cite{ext5, head-analysis, clark-etal-2019-bam, mtl}. We observe a similar trend in the Bert Seq-CLS row of Table \ref{tbl:in-domain-results} and \ref{tbl:out-of-domain-results}, where despite the overarching task of text classification remaining the same when scaling the output space of the classification head to more labels across aspects, we see heavy performance degradation compared to having individual dataset models. For example, in table \ref{tbl:in-domain-results} training a multi-dataset BERT sequence classifier performs worse for every benchmark dataset compared to its single-dataset counterpart. Additionally, for the zero-shot formalizations, we observe the lowest positive transfer on datasets with the lowest level of token overlap between labels seen during training and out-of-domain labels, as shown in Figure \ref{fig:label-overlap}. We theorize that the reason for this phenomenon is that the model is over-fitting to the specific labels seen during training instead of generalizing to the "aspect".

\begin{table*}[t]
    \centering
    \small
    
    \begin{tabular}{ c c c c }

        \toprule \thead{Dataset} & {\thead{Aspect}} & \thead{Train/Test} & \thead{\#labels} \\
        \addlinespace[0.125em] \midrule \addlinespace[0.25em]
        \addlinespace[0.25em] \multicolumn{4}{c}{\textit{in-domain}} \\ \addlinespace[0.25em]
        \hline
        
        \thead{GoEmotions} & sentiment & 43K/5.4K & 28 \\
        \thead{TweetEval} & sentiment & 45K/12K & 3 \\
        \thead{Emotion} & sentiment & 16K/2K & 6 \\
        \addlinespace[0.125em] \hline \addlinespace[0.125em]
        
        \thead{SGD} & intent & 16K/4.2K & 26 \\
        \thead{Clinc-150} & intent & 15K/4.5K & 150 \\
        \thead{SLURP} & intent & 12K/2.6K & 75 \\
        \addlinespace[0.125em] \hline \addlinespace[0.125em]
        
        \thead{AG News} & topic & 120K/7.6K & 4 \\
        \thead{DBpedia} & topic & 560K/70K & 14 \\
        \thead{Yahoo} & topic & 1.4M/60K & 10 \\
        \bottomrule
    \end{tabular} \quad \begin{tabular}{ c c c c }
        \toprule \thead{Dataset} & {\thead{Aspect}} & \thead{Train/Test} & \thead{\#labels} \\
        \addlinespace[0.125em] \midrule \addlinespace[0.25em]
        \addlinespace[0.25em] \multicolumn{4}{c}{\textit{out-of-domain}} \\ \addlinespace[0.25em]
        \hline

        \thead{Amazon Polarity} & sentiment & 3.6M/400K & 2 \\
        \thead{Fin. Phrase Bank} & sentiment & 1.8k/453 & 3\\
        \thead{Yelp} & sentiment & 650K/50K & 3 \\
        \addlinespace[0.125em] \hline \addlinespace[0.125em]
        
        \thead{Banking77} & intent & 10K/3.1K & 77 \\
        \thead{SNIPS} & intent & 14K/697 & 7 \\
        \thead{NLU Eval} & intent & 21K/5.2K & 68 \\
        \addlinespace[0.125em] \hline \addlinespace[0.125em]
        
        \thead{MultiEURLEX} & topic & 55K/5K & 21 \\
        \thead{Patent} & topic & 25K/5K & 9 \\
        \thead{Consumer Finance} & topic & 630K/160K & 18 \\
        \bottomrule
    \end{tabular} 
    \caption{Universal Text Classification Dataset (UTCD) consists of the following datasets: \citet{go-emotions, tweet-eval, emotion, sgd, clinc, slurp, ag-news&yahoo&amazon-polarity&yelp, dbpedia, fin-phrase-bank, banking77, snips, nlu-eval, multi-eurlex, patent, consumer-finance}} 
    \label{tbl:utcd-stats}
\end{table*}

\subsection{Implicit Training}
In order to introduce aspect specification into our zero-shot models, we take inspiration from T5's \cite{t5} text-to-text framework for multi-task generalization. In this framework, the model is fed some text for context and is then asked to produce some output text. As an example, to ask the model to translate the sentence “That is good.” from English to German, the model would be fed the sequence “translate English to German: That is good.” and would be trained to output “Das ist gut.” Similarly, for each aspect (as defined in section \ref{data}), we introduce a conditional aspect token to the model input that acts as a context for that specific aspect. As such, in addition to learning the best contextual representation for the <text, label> input pair, the model implicitly learns a higher level understanding of the underlying aspect. By adding this conditional representation, even as the label space changes, the model is better able to understand the aspect at hand. This is shown in part(b) of figure \ref{fig:train-strategies}. In the case of implicit binary zero-shot classification, the model is additionally provided with a concatenation of the aspect token and the output is selected as:

\begin{equation}
    \hat{y} = \argmax{i \in \{1 \ldots n \}} f(\mathrm{label}(y_i), \mathrm{aspect}(a_{y_i}), x)
\end{equation}

\subsection{Explicit Training}
Given our hypothesis that these language models will be able to generalize to unseen labels as a result of implicitly learning the task at hand, we explore the idea of explicitly training this generalization in a supervised manner. Instead of adding a conditional aspect token, we add an additional pre-training step in which the model is trained on aspect detection. This step acts as an initialization process whereby the model representations are tuned at the aspect level first. Once this step is completed the model is then fine-tuned for its respective zero-shot classification objective. This process is shown in part (c) of figure \ref{fig:train-strategies}. For a given text $x$ this explicit training step is defined as:

\begin{equation}
    \vec{\mathrm{P}} \left( \left\{a_j\right\}_m^1 \big | ~h \right) = \mathrm{softmax}(W h)
\end{equation}

\section{UTCD: Universal Text Classification Dataset} \label{data}
In order to test the zero-shot generalization of these NLP models we introduce UTCD. UTCD is a compilation of 18 classification datasets spanning 3 main aspects of Sentiment, Intent/Dialogue, and Topic classification. A breakdown of each dataset is provided in appendix \ref{apd:dataset}. UTCD focuses on the task of zero-shot text classification where the candidate labels are descriptive of the text being classified. To make NLP models more broadly useful, zero-shot techniques need to be capable of label, domain \& aspect transfer. As such, in the construction of UTCD we enforce the following principles:

\paragraph{Textual labels} In UTCD, we mandate the use of textual labels. While numerical label values are often used in classification tasks, descriptive textual labels such as those present in the datasets across UTCD enable the development of techniques that can leverage the class name which is instrumental in providing zero-shot support. As such, for each of the compiled datasets, labels are standardized such that the labels are descriptive of the text in natural language. 
    
\paragraph{Diverse domains and Sequence lengths} In addition to broad coverage of aspects, UTCD compiles diverse data across several domains such as Banking, Finance, Legal, etc each comprising varied length sequences (long and short). The datasets are listed in Table \ref{tbl:utcd-stats}. 

As described in section \ref{approach}, we define aspect as the sub-task type to which a given set of datasets can belong too. We simulate the Zero-shot learning case by splitting UTCD into \textit{in-domain}, data a given model would be trained on, and \textit{out-of-domain}, data with novel classes unseen during training. Additionally, to prevent data imbalance across aspects, we sub-sample the \textit{in-domain} datasets such that the total number of unique text in each aspect is the same while maintaining class label distribution for each dataset.  Class imbalance is known to degrade performance in deep learning models \cite{BUDA2018249, imbalance}. We observe a similar trend where aspect normalization results in performance improvement.

\begin{table*}[t]
    \centering\small
    \begin{tabular}{@{\extracolsep{2pt}} c c c c c c c c c c c c @{}}
        \addlinespace[0.125em] 
        \toprule \multirow{2}{*}[-1em]{\thead{Model}} & \multirow{2}{*}[-0.75em]{\thead{Training\\Strategy}}
            & \multicolumn{3}{c}{Sentiment} & \multicolumn{3}{c}{Intent} & \multicolumn{3}{c}{Topic} & \multirow{2}{*}[-1em]{\thead{Average}} \\ \addlinespace[0.125em] \cline{3-5} \cline{6-8} \cline{9-11} \addlinespace[0.125em]
            & & \thead{\scriptsize Amazon\\\scriptsize Polarity} & \thead{\scriptsize Fin.\\\scriptsize Phrase Bank} & \thead{\scriptsize Yelp} & \thead{\scriptsize Banking\\\scriptsize 77} & \thead{\scriptsize SNIPS} & \thead{\scriptsize NLU\\\scriptsize Eval} & \thead{\scriptsize Multi\\\scriptsize EURLEX} & \thead{\scriptsize Patent} & \thead{\scriptsize Consumer\\\scriptsize Finance} \\
        \midrule \addlinespace[0.375em]
        
        \multirow{2}{*}[0.25em]{\thead{BERT\\Seq-CLS*}} & \makecell{individual} & 96.0 & 97.2 & 84.8 & 88.6 & 99.0 & 88.9 & 94.8 & 64.1 & 82.6 & 88.4 \\
                                        \addlinespace[0.25em] & \makecell{full} & 93.1 & 24.9 & 79.0 & 84.7 & 97.3 & 87.4 & 81.4 & 50.2 & 76.9 & 75.0 \\
        
        \addlinespace[0.125em] \midrule \addlinespace[0.375em]
        \multirow{3}{*}[0em]{\thead{Binary\\BERT}}
                                                & vanilla & \textbf{80.7} & \textbf{68.9} & 58.5 & \textbf{51.4} & 82.9 & 71.6 & 28.7 & 13.6 & 22.3 & 53.2 \\
                                                & implicit (ours) & 80.1 & 66.0 & \textbf{59.8} & 51.3 & 82.5 & \textbf{73.1} & 30.3 & 15.2 & 23.4 & 53.5 \\
                                                & explicit (ours) & 76.1 & 66.7 & 56.0 & 49.8 & \textbf{83.8} & 69.6 & \textbf{44.5} & \textbf{19.5} & \textbf{30.2} & \textbf{55.1} \\
        
        \addlinespace[0.125em] \midrule \addlinespace[0.375em]
        \multirow{3}{*}[0em]{\thead{Bi-\\Encoder}} 
                                                    & vanilla & 69.9 & \textbf{71.7} & 46.5 & 9.4 & 70.4 & \textbf{71.1} & 33.5 & \textbf{11.7} & 18.4 & 44.7 \\
                                                    & implicit (ours) & \textbf{79.6} & 64.0 & \textbf{56.8} & \textbf{21.1} & \textbf{72.5} & 61.9 & \textbf{35.4} & 9.6 & 11.3 & \textbf{45.8} \\
                                                    & explicit (ours) & 71.5 & 63.6 & 52.1 & 9.7 & 71.9 & 70.0 & 27.4 & 9.3 & \textbf{27.0} & 44.7 \\
        
        \addlinespace[0.125em] \midrule \addlinespace[0.375em]
        \multirow{3}{*}[0em]{\thead{GPT-2\textsuperscript{$\dagger$}}} 
                                                    & vanilla & 88.3 & 71.1 & 70.9 & \textbf{22.8} & 52.2 & 61.7 & 22.3 & 23.5 & 12.6 & 47.3 \\
                                                    & implicit (ours) & 89.3 & 61.4 & \textbf{71.9} & 16.5 & 33.7 & \textbf{63.1} & 18.6 & \textbf{25.8} & 12.2 & 43.6 \\
                                                    & explicit (ours) & \textbf{89.7} & \textbf{75.9} & 71.5 & 22.4 & \textbf{54.1} & 60.7 & \textbf{23.5} & 21.6 & \textbf{13.9} & \textbf{48.2} \\
        
        \addlinespace[0.125em] \midrule \addlinespace[0.25em]
        {\thead{BART\textsuperscript{$\ddagger$}}} & \makecell{Zero-shot} & 91.0 & 40.2 & 75.2 & 42.2 & 61.4 & 40.1 & 19.8 & 8.9 & 24.6 & 44.8 \\
        
        \addlinespace[0.125em] \midrule \addlinespace[0.25em]
        {\thead{GPT-3\textsuperscript{$\ddagger$}}} & \makecell{Zero-shot} & 54.4 & 52.8 & 77.0 & 23.7 & 13.9 & 37.9 & - & - & - & 43.3 \\
        
        \bottomrule
    \end{tabular} \par
    \caption{
        Aspect-Normalized out-of-domain accuracy. 
        *Supervised upper bound, not a zero-shot framework.
        \textsuperscript{$\dagger$}In case none of the given labels are generated at inference, the generated text is embedded and compared with label embeddings.
        \textsuperscript{$\ddagger$}Out-of-the-box zero-shot classifier.
    } 
    \label{tbl:out-of-domain-results}
     \vspace{-1.2pc}
\end{table*}

\section{Experimental Setup}

\paragraph{Model Architectures} \label{architecture}
For binary classification, we use BERT\textsubscript{BASE} with sentence pair classification as in \citet{bert}.
For dual encoding classification, we use Sentence-BERT \citep{sbert} with BERT\textsubscript{BASE} as the base encoder, mean pooling, and cosine similarity as the distance metric. For generative classification, we use the 345M GPT-2 \citep{gpt2} as the language model and the input representation described in \citet{gpt2-nvidia}. These models are denoted Binary BERT, Bi-Encoder, and GPT-2 respectively.

\paragraph{Training} \label{train}
We train all models with AdamW \citep{adamw} and weight decay of 0.01 on all \textit{in-domain} data for 3 epochs, for both pre-training and fine-tuning stages. For explicit pre-training, we use a learning rate of $2\text{e-}5$, batch size of 16, and linear learning rate warmup over the first 10\% steps with a cosine schedule. For binary and dual encoding we use a learning rate of $2\text{e-}5$, batch size of 16, with 10\% warmup and a linear schedule. For generative classification fine-tuning, we use a learning rate of $4\text{e-}5$, batch size of 128, with 1\% warmup and a cosine schedule as reported in \citet{gpt2-nvidia}. We pre-process data and train all models with different random seeds over multiple runs.

\begin{table*}[t]
    \centering\small
    \begin{tabular}{@{\extracolsep{2pt}} c c c c c c c c c c c c @{}}
        \addlinespace[0.125em] 
        \toprule \multirow{2}{*}[-1em]{\thead{Model}} & \multirow{2}{*}[-0.75em]{\thead{Training\\Strategy}}
            & \multicolumn{3}{c}{Sentiment} & \multicolumn{3}{c}{Intent} & \multicolumn{3}{c}{Topic} & \multirow{2}{*}[-1em]{\thead{Average}} \\ \addlinespace[0.125em] \cline{3-5} \cline{6-8} \cline{9-11} \addlinespace[0.125em]
            & & \thead{\scriptsize Go\\\scriptsize Emotions} & \thead{\scriptsize Tweet\\\scriptsize Eval} & \thead{\scriptsize Emotion} & \thead{\scriptsize SGD} & \thead{\scriptsize Clinc\\\scriptsize-150} & \thead{\scriptsize SLURP} & \thead{\scriptsize AG\\\scriptsize News} & \thead{\scriptsize DBpedia} & \thead{\scriptsize Yahoo} \\
        \midrule \addlinespace[0.375em]
        
        \multirow{2}{*}[0.25em]{\thead{BERT\\Seq-CLS*}} & \makecell{individual} & 63.0 & 69.5 & 92.9 & 78.7 & 95.2 & 85.5 & 94.1 & 99.2 & 73.4 & 83.4\\
                                        \addlinespace[0.25em] & \makecell{full} & 56.7 & 55.4 & 91.1 & 80.5 & 82.9 & 77.3 & 86.7 & 98.6 & 66.6 & 77.3 \\
        
        
        \addlinespace[0.125em] \midrule \addlinespace[0.375em]
        \multirow{3}{*}[0em]{\thead{Binary\\BERT}}
                                                & vanilla & 59.3 & \textbf{67.6} & \textbf{92.4} & 91.5 & 87.8 & \textbf{81.8} & \textbf{90.0} & 98.9 & 67.9 & 81.9 \\
                                                & implicit (ours) & 59.9 & 67.2 & 91.8 & \textbf{93.5} & 87.1 & 81.8 & 89.2 & \textbf{98.9} & \textbf{68.1} & \textbf{82.0} \\
                                                & explicit (ours) & \textbf{60.2} & 66.6 & 91.8 & 93.4 & \textbf{88.0} & 80.4 & 88.7 & 98.9 & 67.8 & 81.7 \\
        
        \addlinespace[0.125em] \midrule \addlinespace[0.375em]
        \multirow{3}{*}[0em]{\thead{Bi-\\Encoder}} 
                                                    & vanilla & \textbf{59.2} & 65.7 & 92.8 & 82.2 & \textbf{84.4} & 79.9 & 89.3 & 99.0 & 67.4 & 80.0 \\
                                                    & implicit (ours) & 56.9\ & 66.0 & 90.9 & \textbf{81.3} & 82 & 78.9 & 88.8 & \textbf{99.0} & \textbf{67.9} & 79.1 \\
                                                    & explicit (ours) & 58.8 & \textbf{66.8} & \textbf{91.8} & 82.7 & 83.3 & \textbf{79.9} & \textbf{89.5} & 98.9 & 67.7 & \textbf{80.0} \\
        
        \addlinespace[0.125em] \midrule \addlinespace[0.375em]
        \multirow{3}{*}[0em]{\thead{GPT-2\textsuperscript{$\dagger$}}} 
                                                    & vanilla & 58.8 & \textbf{70.6} & 75.9 & 84.2 & 81.4 & 75.3 & 86.7 & 98.5 & 68.3 & 77.7 \\
                                                    & implicit (ours) & 59.0 & 70.3 & 71.4 & \textbf{84.7} & 81.7 & 73.1 & 87.7 & 98.4 & 68.3 & 77.2 \\
                                                    & explicit (ours) & \textbf{60.1} & 70.1 & \textbf{76.4} & 84.3 & \textbf{81.9} & \textbf{76.7} & \textbf{87.9} & \textbf{98.6} & \textbf{68.6} & \textbf{78.3} \\
        
        \addlinespace[0.125em] \midrule \addlinespace[0.375em]
        {\thead{BART\textsuperscript{$\ddagger$}}} & \makecell{Zero-shot} & 24.2 & 47.8 & 37.7 & 41.4 & 50.4 & 27.5 & 71.7 & 65 & 49.2 & 46.1 \\
        
        \bottomrule
    \end{tabular} \par
    \caption{
        Aspect-Normalized in-domain accuracy. 
    } 
    \label{tbl:in-domain-results}
    \vspace{-1.2pc}
\end{table*}

\section{Results \& Discussion} \label{results}

In this section we present and analyze the results of our experiments, detailing our insights and discussing the implications of each of our techniques.

\paragraph{Evaluation Task} \label{eval}
We report accuracy on the test set of all \textit{in-domain} and \textit{out-of-domain} datasets. In multi-label cases where there is more than one valid label, the prediction is considered correct if the model predicts any one of the correct labels. 
For generative classification, we observe instances in which GPT-2 may not generate one of the label options, a known problem for PLM generation \citep{gpt, control-gen}. In such cases, we consider the label option most similar to the generated answer as prediction, by mapping the generated output and the valid classes to an embedding space. For this encoding, we use the pre-trained model MPNet \citep{mpnet} with mean pooling encoder from Sentence-BERT \citep{sbert} for mapping the labels and cosine similarity as the distance metric. This ensures the consistency of GPT-2's output with the other zero-shot formalizations.

\paragraph{Upper-bound \& Zero-shot Baselines} To gauge the ability of our models to generalize to unseen data, we establish our upper-bound as the performance of a fully supervised model on the target data. Specifically, we fine-tune two variations of BERT\textsubscript{BASE} for sequence classification which we denote as \textit{"individual"} and \textit{"full"}. For \textit{individual}, we fine-tune a dedicated classification model for each dataset in UTCD. For \textit{full}, we fine-tune a single model for all datasets. Additionally, we compare the zero-shot performance of our models to the popular LLM GPT-3 \cite{gpt3}, and BART MNLI \cite{benchmark&bert-nli} which is the most popular and widely downloaded zero-shot model on Huggingface Hub\footnote{\url{https://huggingface.co/facebook/bart-large-mnli}}.

\subsection{Out-of-domain Performance} \label{results:out}
In table \ref{tbl:out-of-domain-results}, we report results on the out-of-domain test set for UTCD. To evaluate the ability of our zero-shot models to adapt to unseen data, we evaluate our fine-tuned models from table \ref{tbl:in-domain-results} on the out-of-domain test set without training on any out-of-domain data. Across the zero-shot formalizations, we observe that our explicit Binary BERT achieves the best performance with a 2\% increase over its vanilla counterpart. Thus showing the power of the explicit pre-training strategy for binary classification formalization.

When compared to the "full" supervised out-of-domain model, despite having not been trained on any data from the target dataset, across the aspects of sentiment and intent, our models are able to generalize well. Specifically, across all formalizations, our models are able to outperform the supervised model on the financial phrase bank dataset. We observe that this drop is due to conflicting domain data. UTCD's out-of-domain set consists of similar financial datasets in the other aspects of intent and topic. Given that examples from the finance phrase banks dataset are general in nature, without seeing the label, it is difficult for the sequence classifier to understand the task at hand, thus causing it to classify to conflicting labels from similar datasets. This showcases the need to include aspect-specific knowledge.

Lastly, when inspecting the performance of vanilla fine-tuning compared implicit and explicit training, we are able to outperform vanilla on generalizing to unseen data on 6, 6, and 8 of the 9 datasets in out-of-domain UTCD across Binary BERT, Bi-encoder, and GPT-2 models respectively. In particular, for explicit training on Binary BERT, we achieve a massive improvement in zero-shot generalization (as much as +\%16 for the topic aspect, +9\% on average). Additionally, in comparison to the massive zero-shot baselines of BART and GPT-3 our models are able to outperform on 7 and 8 of the 9 datasets respectively.

\subsection{In-domain Performance} \label{results:in}
In table \ref{tbl:in-domain-results}, we report results on the in-domain test set for UTCD. For in-domain, we conduct implicit \& explicit training across each zero-shot formalization. We observe that when compared with the "full" supervised model, our zero-shot models are more performant while maintaining the flexibility of facilitating zero-shot. When compared with the "individual" variation, as our zero-shot models are trained jointly across different datasets, we achieve better performance than the single supervised model on datasets such as SGD, showing the power of knowledge transfer from other intent datasets such as Clinc-150 \& SLURP.

For vanilla fine-tuning without implicit or explicit training, we observe that across zero-shot formalizations, injecting task specification through implicit and explicit pre-training preserves performance for in-domain data. Showing that while achieving better zero-shot transfer ability our models do not suffer performance loss on data already seen during training.

\subsection{Importance of Label token overlap} 
\begin{figure}[t]
    \centering
    \includegraphics[width=\columnwidth]{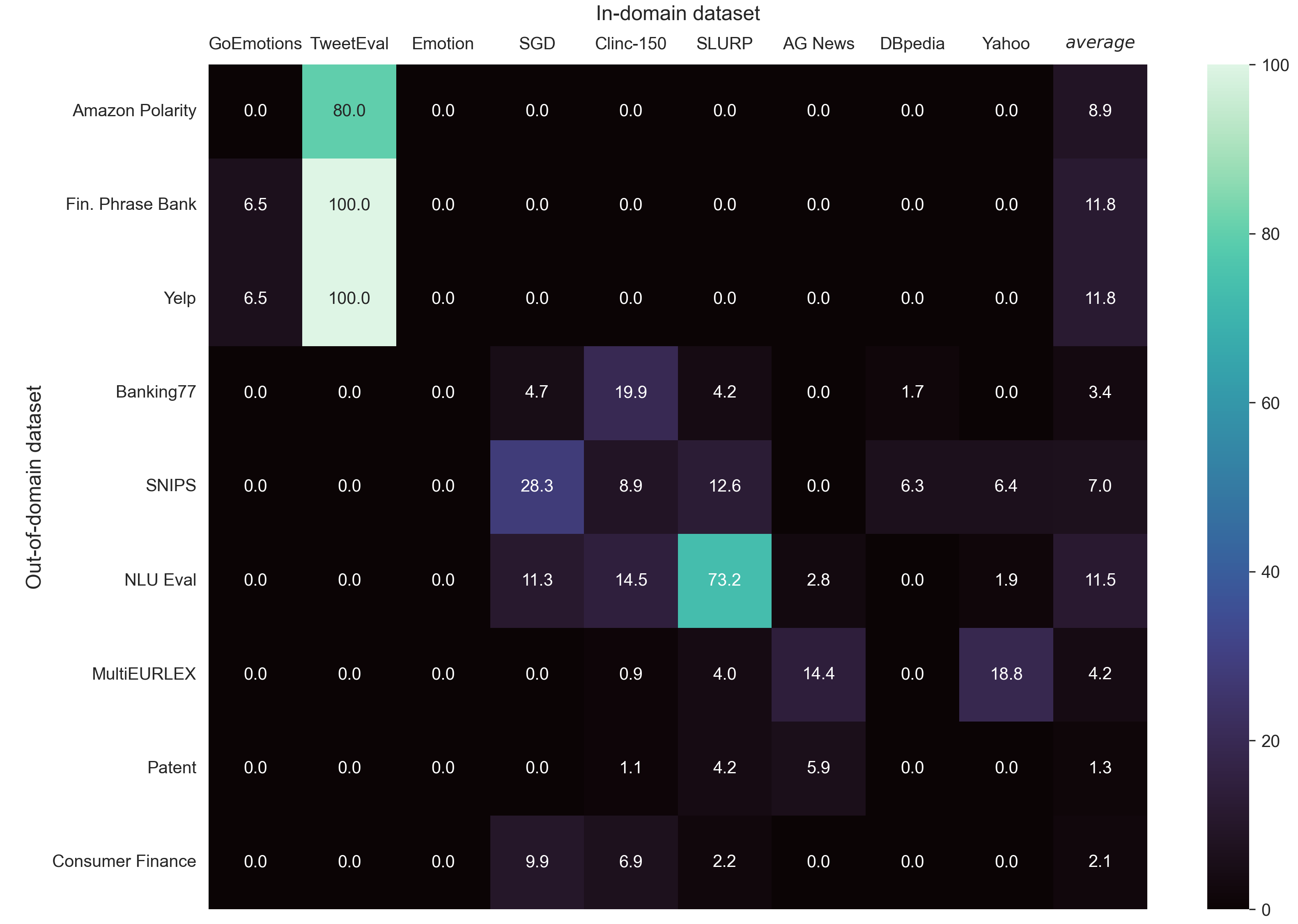}
    \caption{UTCD Out-of-domain Dataset Label Pair-wise Overlap with In-domain Dataset. 0 is no overlap, 100 is exactly the same label set. From sentiment to intent to topic, label overlap decreases in general.}
    \label{fig:label-overlap}
     
\end{figure}

In addition to the need for aspect-specific knowledge, we also observe a high correlation in zero-shot generalization results between the overlap of tokens seen during training and those evaluated on the out-of-domain test. Figure \ref{fig:label-overlap} shows the pairwise overlap of label tokens across the in-domain and out-of-domain datasets. When inspected across aspects, we see that our models are able to achieve the best out-of-domain performance on datasets with the most overlapping label tokens to those seen during training.

\section{Related Work} \label{related_work}
Zero-shot text classification is the task of classifying text into novel categories unseen during training. Early zero-shot classification studies frame the problem as binary classification on whether a given label describes the text \citep{train-once,benchmark&bert-nli}. With the advancement of PLMs, subsequent works \citep{benchmark&bert-nli, gpt2-nvidia} rely on transformer architectures to learn representations from descriptive labels passed in. In particular, \citet{gpt2-nvidia} fine-tune an autoregressive language model to generate titles based on a prompt template containing Tweet articles and a list of title options. Though the model is trained on a great variety of title options, the approach limits the learning to topic classification only, as the authors only analyze performance on topic datasets, unlike our approach which considers a wide array of aspects, each requiring focus on different sections of a given text. 

\citet{benchmark&bert-nli} similarly categorize zero-shot text classification by aspects and implicitly introduce aspects during training with a dedicated template for each aspect. They further propose the classification of a text, label pair as a logic entailment problem. However, the authors analyze a less challenging zero-shot case where a model is trained on a subset of text, label pairs, and evaluated on the remaining text with unseen labels in the same domain. 
Additionally, the authors introduce WordNet definition of the labels as the labels are all single words. This process requires manual intervention and is not applicable for multiple-word label sequences common in intent classification, such as "Check Balance". 
Our work evaluates a more diverse set of datasets for each aspect and a more comprehensive set of zero-shot architectures. 



\section{Conclusion}
In this paper, we investigate the task of zero-shot text classification with the aim of improving the ability of PLMs to generalize both seen and unseen data across domains without the need for additional training. We introduce two new simple yet effective pre-training strategies, \textit{Implicit training} \& \textit{Explicit pre-training} which specifically inject aspect-level understanding into the model at train time. To evaluate this, we release UTCD, a new benchmark dataset for evaluating text classification in zero-shot settings. Experimental results on UTCD show that our approach achieves improved zero-shot generalization on a suite of challenging datasets in UTCD and across many zero-shot formalizations.

\section{Limitations}
While our approach is shown to be effective in improving the zero-shot adaption ability of these PLMs, the scope of this work has only been extended to English languages and has not been tested on other languages. In addition, another limitation of this work is the scope of the aspect. Aspect is defined across 3 main categories of intent, sentiment, and topic in the work. However, given the massive space of text label interpretations, our aspect range can be refined and expanded even further, lending to more analysis of the stability of implicit \& explicit training as the number of aspects grows. We do not investigate this scenario in this work.

\section*{Acknowledgements}
We thank our anonymous reviewers for their feedback and suggestions. This work is supported in part by award NSF1539011 by the National Science Foundation.

\bibliography{custom}
\bibliographystyle{acl_natbib}

\appendix
\label{apd}

\section{UTCD Datasets} \label{apd:dataset}

UTCD is a compilation of 18 classification datasets spanning 3 categories of Sentiment, Intent/Dialogue and Topic classification. UTCD focuses on the task of zero-shot text classification where the candidate labels are descriptive of the text being classified. UTCD consists of ~ 6M/800K train/test examples. 

For sentiment we have the datasets Go Emotion \cite{go-emotions}, TweetEval \cite{tweet-eval}, Emotion \cite{emotion}, Amazon Polarity \cite{ag-news&yahoo&amazon-polarity&yelp}, Finance Phrasebank \cite{fin-phrase-bank} and Yelp \cite{ag-news&yahoo&amazon-polarity&yelp}. The GoEmotions dataset contains 58k carefully curated Reddit comments labeled for 27 emotion categories or Neutral. The TweetEval dataset consists of seven heterogenous tasks in Twitter, all framed as multi-class tweet classification. The tasks include - irony, hate, offensive, stance, emoji, emotion, and sentiment. We used the sentiment portion of this dataset for UTCD. Emotion is a dataset of English Twitter messages with six basic emotions: anger, fear, joy, love, sadness, and surprise. The Amazon Polarity dataset consists of reviews from Amazon. The data spans a period of 18 years, including ~35 million reviews up to March 2013. Reviews include product and user information, ratings, and a plaintext review. The Finance Phrasebank dataset consists of 4840 sentences from English language financial news categorised by sentiment. The Yelp dataset consists of over 600k reviews for the task of sentiment classification.

For the intent/dialogue aspect we have the datasets: Schema Guided Dialgoue \cite{sgd} is an annotated multi-domain, task-oriented conversations between a human and a virtual assistant. Clinc-150 \cite{clinc} is an intent classification (text classification) dataset consisting of 150 in-domain intent classes. SLURP \cite{slurp} is dialuge dataset derived from SLU systems English spanning 18 domains. Banking77 \cite{banking77} is an intent classification dataset for the banking domain. It comprises 13,083 customer service queries labeled with 77 intents. Snips is an NLU dataset of over 16,000 crowdsourced queries distributed among 7 user intents. NLU Evaluation \cite{nlu-eval} is an NLU dataset from the conversational domain annotated with corresponding intents and dialogue scenarios.

Lastly, for the topic aspect we have the datasets: AG News \cite{ag-news&yahoo&amazon-polarity&yelp} is a topic classification dataset extract from the AG News article corpus. It consist of 4 classes from the original corpus. Each class contains 30,000 training samples and 1,900 testing samples. Yahoo Answers dataset \cite{ag-news&yahoo&amazon-polarity&yelp} contains 4,483,032 questions and their answers across 10 categories. Each class contains 140,000 training samples and 5,000 testing samples. DBpedia \cite{dbpedia} dataset is a topic classification dataset constructed from picking 14 non-overlapping classes from DBpedia 2014. Multi Eurlex \cite{multi-eurlex} is a multilingual dataset for topic classification of legal documents. The dataset comprises 65k European Union (EU) laws, officially translated in 23 languages, annotated with multiple labels from the EUROVOC taxonomy. Big Patent \cite{patent} is a topic classification dataset for the legal domain consisting of 1.3 million records of U.S. patent documents along with human written abstractive summaries. Consumer Finance \cite{consumer-finance} dataset is a collection of complaints about consumer financial products and services sent to companies for response.

\end{document}